# Qualitative Order of Magnitude Energy-Flow-Based Failure Modes and Effects Analysis

**Neal Snooke**                                                      NNS@ABER.AC.UK
*Department of Computer Science, Aberystwyth University,*
*Penglais, Aberystwyth, Ceredigion, SY23 3DB, U.K.*

**Mark Lee**                                                         MHL@ABER.AC.UK
*Department of Computer Science, Aberystwyth University.*

## Abstract

This paper presents a structured power and energy-flow-based qualitative modelling approach that is applicable to a variety of system types including electrical and fluid flow. The modelling is split into two parts. Power flow is a global phenomenon and is therefore naturally represented and analysed by a network comprised of the relevant structural elements from the components of a system. The power flow analysis is a platform for higher-level behaviour prediction of energy related aspects using local component behaviour models to capture a state-based representation with a global time. The primary application is Failure Modes and Effects Analysis (FMEA) and a form of exaggeration reasoning is used, combined with an order of magnitude representation to derive the worst case failure modes.

The novel aspects of the work are an order of magnitude(OM) qualitative network analyser to represent any power domain and topology, including multiple power sources, a feature that was not required for earlier specialised electrical versions of the approach. Secondly, the representation of generalised energy related behaviour as state-based local models is presented as a modelling strategy that can be more vivid and intuitive for a range of topologically complex applications than qualitative equation-based representations. The two-level modelling strategy allows the broad system behaviour coverage of qualitative simulation to be exploited for the FMEA task, while limiting the difficulties of qualitative ambiguity explanation that can arise from abstracted numerical models. We have used the method to support an automated FMEA system with examples of an aircraft fuel system and domestic a heating system discussed in this paper.

## 1. Introduction

Qualitative representations (QR) and reasoning have a number of well documented advantages for Failure Modes and Effects Analysis. Three of the most important are that analysis can be performed early in the design life cycle, that there is broad coverage of system behaviour and faults, and that the results are at a level of abstraction which readily maps to system functional states.

QR has been widely applied to electrical systems (de Kleer, 1984; Mauss & Neumann, 1996; Price, Snooke, & Lewis, 2006; Flores & Farley, 1999) and used for a variety of design analysis applications including design concept analysis (gaining an overview of system behaviour), diagnosis, FMEA safety analysis, incremental and multiple fault FMEA analysis (Price & Taylor, 1997), fault tree analysis (FTA) (Price, Wilson, Timmis, & Cain, 1996), and sneak circuit analysis (Price, Snooke, & Landry, 1996; Savakoor, Bowles, & Bonnell, 1993). The qualitative information is an enabling technique allowing significant system nominal and failure states to be identified and distinguished easily and efficiently.

The qualitative results have limitations, the most significant being behavioural ambiguity where alternative behaviours are predicated by the simulation. While often seen as a problem of qualitative techniques, such ambiguity can be turned into an advantage, provided they predict real physical behaviours since they indicate where key design parameters exist. One of the benefits of the modelling structure developed in this paper is that many such ambiguities are constrained and therefore enhance design knowledge rather than being an analysis limitation.

The typical approaches to qualitative modelling and reasoning use qualitative versions of numerical equations derived from component models. While this forms a sound platform, it can be burdensome to extract the relevant from the unimportant with standard equations (e.g fluid flow), and deal with landmarks, integrals or deviations etc. Given the appeal of QR for abstracted explanation and its ability to provide broad analysis of system state, we propose a more specialised view that can be applied to a wide range of engineering domains based on the relationships between generalised physical variables shown in Figure 1.

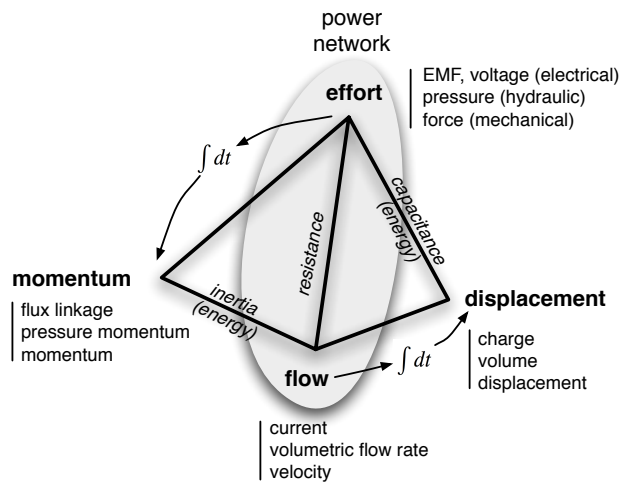

Figure 1: Tetrahedron of state

This paper provides a consistent framework for several earlier implementations, and develops a supporting qualitative order of magnitude (OM) capability (Section 2.1) throughout the models and simulation. The OM representation is derived from a many valued resistance representation (Lee, 2000b) and although the many valued concept and its modelling and reasoning benefits are completely applicable to the OM representation, the actual technique for solving the circuits in that work was based on path labelling (Lee, 1999), and was not a general solution applicable to any network topology. The present paper provides a general solution and extends the technique from purely electrical systems into other domains. As a necessary part of this effort an energy-flow-based formalisation is adopted, comprising a global instantaneous power network (Section 2) and local component models that include the notions of energy and time (Section 3).

Previous work by the authors (Snooke, 2007) on the modelling of fluid flow systems was based on an electrical network analyser (Lee, 1999) but could only deal with series parallel reducible circuits, and this becomes a limitation for topologically complex systems. The incorporation of equivalent resistance reduction using star–delta (Y-$\Delta$) transforms (Mauss & Neumann, 1996) into the OM representation (Section 2.2) allows effort and flow variables for resistive networks of any topology to be solved by reduction and flow assignment expansion (Section 2.3), and completes the work started by Lee. Complex network topologies are due in part to the need to represent concepts such as the 'atmosphere' in fluid flow systems, to allow failure modes such as leaks in closed systems, as well as vented elements in other systems. This unique 'zero' node is common to several domains and is generalised in the present work in Section 2.4.

Applying the techniques derived by Lee (1999) to non electrical systems reveals two limitations. Firstly, the analysis supports only a single effort source in the network, and secondly there is no concept of the substance flowing in the network. Section 2.5 extends the analysis to include multiple power sources by applying the principle of superposition to allow any system to be decomposed into multiple resistive networks. Section 2.6 considers the representation of substances for networks where behaviour is dependent upon the substance carrying the effort and flow. The final part of section 2 deals with some common special cases that can simplify the analysis and maintain vividness of the representation by avoiding unnecessary (Y-$\Delta$) transforms.

Section 3 of the paper is devoted to the local component concept that captures temporal information and thus models the displacement and momentum aspects of Figure 1. The OM modelling of time is the subject of section 3.1 and underlies a Finite State Machine (FSM) based modelling approach. This representation is an improvement on a more restricted predecessor used for electrical component behaviour (Snooke, 1999), which was the foundation for a variety of automated electrical design analysis techniques. Sections 3.2 and 3.3 provide concrete examples of component modelling and system modelling respectively.

Section 4 considers the use of the OM representation to perform exaggeration reasoning for failure analysis and section 5 provides a brief overview of the strategy (Price, 1998; Lee, Bell, & Coghill, 2001) used to process multiple simulation results into an FMEA output. We provide two case study examples in sections 6 and 7. The first case study illustrates the behaviour simulation for a fluid flow system with interesting characteristics, and the second outlines the use of the simulation results to automatically produce a completed FMEA report for an industrial system supplied by the sponsors of the work.

In this paper we utilise a two-level modelling strategy. The lower level (centre section of Figure 1) provides a network-based global qualitative solver for linear resistive networks using graph-based methods that determine instantaneous power (effort and flow) in a network. The higher level utilises the results of the lower level network analysis to decide localized component behaviour from state-based models with a qualitative representation of time as the only global parameter.

This two-level strategy gives significant benefits over other approaches to modelling the kinds of electrical and hydraulic systems dealt with here. The two-level modelling allows separation of the inherently global, instantaneous (single state) power variables, and the local energy-based behaviour of components. This separation is particularly suitable for qualitative representation because qualitative behaviour is fundamentally state-based; any qualitative simulation will produce a sequence of states of the power flows in a system's components. Power flow is a global phenomenon where cause and effect is undetermined and can be very efficiently analysed as a network, avoiding problems of causal instability and ambiguity arising from systems of equations generated from interacting components with local behaviours (Skorstad, 1992a). Many QR approaches use a local propagation method connecting the behaviours of components and have the appeal of generality, however, they often require some way of including additional global information such as equivalent circuits (Sussman & Steele Jr, 1980), or mechanisms for propagating connectivity information (Struss, Malik, & Sachenbacher, 1995). Our approach allows users to build reusable models of components that can be placed in a library and used just by describing how components are linked together.

Bond graphs share the same underlying energy-flow-based concept as the lower level of our approach, and are a popular domain-independent graphical method for describing a system first described by Paynter (1961) and developed by numerous others. They are an excellent way of generating the systems of differential equations that describe system behaviour. Qualitative versions of bond graphs can be produced (Ghiaus, 1999), however these are qualitative versions of the standard numerical equations and therefore suffer the same difficulties as any of the general equation-based constraint methods. A qualitative version of bond graphs is successfully used to model energy flows associated with people in a building (Tsai & Gero, 2010), where the movement of people is not determined by any specific physical laws and thus there is no choice but to provide an ad-hoc model based on high-level knowledge about the building use. It has been observed that qualitative models often result in too many fault candidates, and a reasoning scheme based on past experience and ad-hoc system dynamics where it can be difficult to obtain the proper models for quantitative methods (Samantaray & Ould, 2011). The most widely-used qualitative equation-based constraint method has been QSIM (Kuipers, 1986). Kuipers has acknowledged the problems with QSIM for many applications - chattering, uncontrolled branching of possibilities, asymptotic behaviours (Fouché & Kuipers, 1990). While some techniques have been developed for minimising the problems (Clancy & Kuipers, 1997), the problems with the approach remain for practical systems. As observed (Mosterman & Biswas, 2000), reducing model complexity by eliminating higher-order derivatives and non-linear effects leads to discontinuities in the system (i.e. state changes) and careful analysis of the underlying physical nature of the system is required when constructing models in order to ensure that the simplified models correspond to real behaviour. That kind of careful analysis is not possible when linking QSIM-style qualitative equation-based constraint methods to physical components, and certainly not when wanting to be able to simulate faulty components.

Our two-level approach has achieved practical success even in less sophisticated versions than the present approach (Price & Struss, 2004) by avoiding direct translation of differential equations to their qualitative versions, and rather representing phenomena explicitly as state-based representations at the local component behaviour level and providing a constrained global representation. This does not limit the applicability because it is the property of the model that is discrete or continuous rather than any inherent property of the system (Struss, 2003), and the real question is what kind of model is appropriate for the reasoning task at hand. The modelling in this paper requires minimal algebraic effort, by representing component behaviour at an abstract level such that qualitative behaviour can be predicted for nominal and failure modes for topologically complex systems with many component states.

| Effort, $E$ | Resistance, $R$ | Flow, $F$ |
|---|---|---|
| 0 | 0 | ? |
| 0 | $r$ | 0 |
| 0 | $\infty$ | 0 |
| $u$ | 0 | $\sqcup$ |
| $u$ | $r$ | $f$ |
| $u$ | $\infty$ | 0 |

Table 1: Qualitative electrical current assignment

## 2. The Qualitative Power Network

The power network (system circuit) is represented using resistances, $R$, as the structural component. The simulation task is to derive the effort across each resistance and the flow through each resistance given a power (effort or flow) source in the network. A minimal useful qualitative quantity space uses $[0, r, \infty]$ to represent resistance and $[0, u]$ to represent effort, $E$, at the supply terminals (Lee, 1999). A linear network uses the generalised version of Ohm's Law for effort, flow, and resistance, $E = F \times R$, and provides the current assignment in Table 1. The first row is qualitatively ambiguous because any flow through a zero resistance produces no effort loss. It is not physically possible to have an effort drop across a zero resistance and hence the flow in the fourth row is shown as $\sqcup$ to indicate an impossible case.

A power network is considered as a graph $G(T, A)$ containing nodes $T$ and edges $A$ that connect exactly two nodes, and represent the circuit resistances. Each edge $e \in A$ has a resistance value $R(e)$ and connects a pair of nodes $e = \langle t_1 \in T, t_2 \in T \rangle$. Effort is measured between two nodes $E(t_1, t_2)$ and flow is measured through an edge $F(e)$. The *degree* of a node is the number of connections to that node. Looped edges with both ends connected to the same node therefore increase the degree of the node by two.

### 2.1 Order of Magnitude Representation

An order of magnitude representation for $R$ allows more detailed modelling without introducing qualitative ambiguity by separating artefacts with significantly different characteristics (Raiman, 1991; Lee, 2000a). This enhancement improves the ability to represent nominal behaviour such as distinguishing signal level power from actuator power in electrical circuits, and gravitational 'head' pressure from the system pump in selected fluid transfer systems. The modelling of faults is also improved by allowing exaggerated faults to be modelled, thus producing effects for faults where effects would otherwise be qualitatively indistinguishable from nominal operation.

We define $O(M)$ qualitative resistance values $R = [r_1 = 0, r_2, \ldots, r_{i-1}, r_m = \infty]$ such that $r_{i+1} \gg r_i$ for any $i$. In addition $r_i / r_{i+1} = r_j / r_{j+1}$ for any $i, j \in \mathbb{N}$. Physically we interpret this to mean that any number of $r_m$ valued resistors in series will be dominated by a single $r_{m+1}$ valued resistor in the same series segment, and that in addition, the magnitudes of the resistances are qualitatively of equal spacing. The OM representation proposed has the benefit that it does not generate additional qualitative landmarks that lead to potential ambiguity in the model, indeed its purpose is to allow qualitative distinctions for characteristics that are known to be of a significantly different magnitude. The qualitative OM approach is somewhat analogous to the base 10 logarithmic scale used for magnitude approximation in numerical reasoning, with two differences. For the qualitative $\gg$ assumption above to hold generally when mapping from numerical specifications to qualitative ones in practical systems, a coarser magnitude scale than 10 is required. Therefore, models that distinguish values based on the numerical equivalent of three order of magnitude prefix ranges, such as A, mA, $\mu$A for electrical current, work well. Secondly, $\gg$ may be interpreted for a specific application type or domain. For example if we can assume in some fluid system that the pressure in a system created by gravitational 'head' in a vertical pipe is always dominated by the pump(s) working against it, regardless of the number of pipe sections in the system, we can consider these at different qualitative magnitudes, even though numerically this clearly is not the case. In this case a consistent set of qualitative magnitudes, models, and interpretations is required for each application or domain that separates phenomena considered to satisfy $\gg$ for target systems.

| $a$ | $b$ | $a \times b$ |
|---|---|---|
| 0 | 0 | 0 |
| 0 | $r^{\rhd n}$ | 0 |
| 0 | $\infty$ | ? |
| $r^{\rhd n}$ | $r^{\rhd m}$ | $r^{\rhd(n+m)}$ |
| $r^{\rhd n}$ | $\infty$ | $\infty$ |
| $\infty$ | $\infty$ | $\infty$ |

Table 2: OM Multiplication

The qualitative magnitude notation for $q^{\rhd n}$ is introduced as a convenience for lower magnitudes to indicate a quantity $n$ orders of magnitude below $q$. Similarly, for convenience, $q^{\lhd n}$ indicates a quantity $n$ orders of magnitude above $q$ and therefore $q^{\rhd n} = q^{\lhd(-n)}$. The OM calculations follow the usual rules of sign algebra (Travé-Massuyès, Ironi, & Dague, 2004) but allows the domination of one magnitude by another as shown in Tables 2 and 3 for positive multiplication and addition respectively. Table 1 can now have $u$ replaced by $u^{\rhd n}$ and $r$ replaced by $r^{\rhd m}$ resulting in $f^{\rhd(n-m)}$ for the flow result in row 5.

## 2.2 Network Reduction

The OM representation is utilised for each of the resistance, effort, and flow variables of a power network. The aim is to derive flow and effort values throughout the entire circuit, and this is achieved by reducing the circuit to a single equivalent resistance, assigning the flow value and expanding the network assigning effort and flow values at each level of expansion according to the qualitative structures present. The network reduction is performed by series and parallel (SP) circuit simplification with an OM version of the qualitative star–delta (Y-$\Delta$) transformation (Mauss & Neumann, 1996) for non SP cases as follows.

| + | 0 | $+p^{\triangleright n}$ | $-p^{\triangleright n}$ | $\infty$ | ? |
|---|---|---|---|---|---|
| 0 | 0 | $+p^{\triangleright n}$ | $-p^{\triangleright n}$ | $\infty$ | ? |
| $+p^{\triangleright m}$ | $+p^{\triangleright m}$ | $+p^{\triangleright \min(m,n)}$ | $-p^{\triangleright n}$ if $n<m$ <br> $+p^{\triangleright m}$ if $m<n$ <br> ? if $n=m$ | $\infty$ | ? |
| $-p^{\triangleright m}$ | $-p^{\triangleright m}$ | $+p^{\triangleright n}$ if $n<m$ <br> $-p^{\triangleright m}$ if $m<n$ <br> ? if $n=m$ | $-p^{\triangleright \min(m,n)}$ | $\infty$ | ? |
| $\infty$ | $\infty$ | $\infty$ | $\infty$ | $\infty$ | ? |
| ? | ? | ? | ? | ? | ? |

Table 3: OM addition

Within a network graph G, any nodes of degree two can be removed and have both connected edges $e_1, e_2$ replaced with an equivalent series edge of resistance:

$$R(e_1 \mathbin{\vdots} e_2) = \max(R(e_1), R(e_2)) \tag{1}$$

Any edges $e_1, e_2, \ldots, e_n$ that share the same pair of nodes can be replaced by a single equivalent parallel edge with the resistance:

$$R(e_1 \| e_2 \| \ldots \| e_n) = \min(R(e_1), R(e_2), \ldots, R(e_i)) \tag{2}$$

To facilitate vividness of the derived model we use the convention $e_1 \| e_2$ to label edges representing a parallel combination of edges and $e_1 \mathbin{\vdots} e_2$ to indicate a series combination. Iterative application will produce a tree of equivalent resistances and will result either in a single resistance $R'$ between the supply nodes $t_\oplus$ and $t_\odot$ or a non-SP reducible circuit fragment.

The majority of circuits are SP reducible, particularly if we consider that zero resistance edges can be removed from the network to simplify the topology (Section 2.7). For the remainder a Y-$\Delta$ transformation can be applied as shown in Figure 2. The introduction of Y-$\Delta$ resistances unfortunately reduces the vividness of the representation since they are not directly related to the original component structure; however, it provides a general solution and can assign flow direction for any network, unlike earlier work (Lee, 2000a). A qualitative signs version of this transformation was utilised before (Mauss & Neumann, 1996), however, an OM representation introduces the possibility of additional levels of resistance for the transformed node and requires a more detailed analysis.

Edges $e_1 \ldots e_n$ connected to a non SP reducible degree $n$ star node are replaced by new edges $e_{jk}$ that form an equivalent network where $1 \leq j \leq n, 1 \leq k \leq n, k > j$. Using $R^{\mathbb{R}}(e)$ to notate the numerical resistance of edge $e$, a 3 node Y-$\Delta$ transformation is defined (symmetrically for other edges) as:

$$R^{\mathbb{R}}(e_{12}) = \frac{R^{\mathbb{R}}(e_1)R^{\mathbb{R}}(e_2) + R^{\mathbb{R}}(e_1)R(e_3) + R^{\mathbb{R}}(e_2)R^{\mathbb{R}}(e_3)}{R^{\mathbb{R}}(e_3)} \tag{3}$$

Equation 3 is generalised as a star mesh transform defined for star nodes of any order as:

$$R^{\mathbb{R}}(e_{jk}) = R^{\mathbb{R}}(e_j)R^{\mathbb{R}}(e_k) \sum_{m=1}^{n} 1/R^{\mathbb{R}}(e_m) \tag{4}$$

Noting that for the OM qualitative values:

$$max\left(\frac{1}{a}, \frac{1}{b}, ...\right) = \frac{1}{\min(a, b, ...)} \tag{5}$$

the qualitative version of equation 4 for $1 \leq m \leq n$ is:

$$R(e_{jk}) = \frac{R(e_j)R(e_k)}{\min(R(e_m))} \tag{6}$$

If the denominator is $\infty$, then all of the resistances must be $\infty$ and $R(e_{jk}) = \infty$. If any of the star resistances, $R(e_m) = 0$, then the result is undefined. Hence, for efficiency, zero resistance edges are removed by combining the associated nodes into a *super node* which will be discussed in the next subsection. If the numerator contains 0 or $\infty$, that value is the result. For all other cases the result is determined by the magnitude indices of the values. If $min(R(e_m)) = r_m^{\lhd c}$ and $R(e_j) = r_j^{\lhd a}$ and $R(e_k) = r_j^{\lhd b}$,

$$r_{jk}^{\lhd(a+b-c)} = \frac{r_j^{\lhd a} r_k^{\lhd b}}{r_m^{\lhd c}} \tag{7}$$

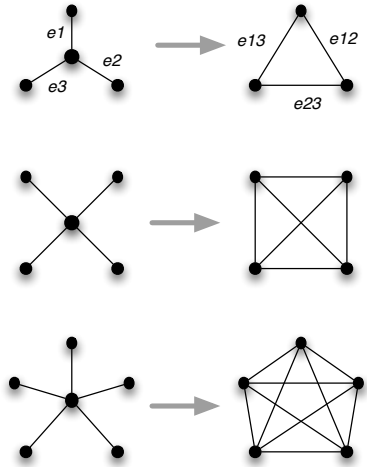

Figure 2: Δ-Y transformation for 3, 4 and 5 nodes

## 2.3 Flow and Effort Assignment

The reduced network can have a flow (or effort for flow sources) assigned for its source directly from the OM extended version of Table 1. The $R = 0$ case ⊔ is reported immediately since it has a specific physical interpretation in most circumstances, such as an electrical short circuit. When $R = 0$ contains another source (Section 2.5) not presently under consideration, that edge is ignored. To determine the flow in a specific circuit component, the edge hierarchy is expanded for the required edge. The sign of a flow determines its direction relative to an arbitrary terminal order for each resistor, using a convention $t_1 \to t_2$ is $f$ and $t_2 \to t_1$ is -$f$.

For edges $e_1, e_2$ that are part of a series pair the flow through each is simply the flow through the pair: $F(e_1) = F(e_2) = F(e_1 \,\vdots\, e_2)$ and the effort across the edge is $E(e_1) = F(e_1)R(e_1)$ (noting that $f^{\triangleright n} r^{\triangleright m} = u^{\triangleright(m+n)}$). If $R(e) = \infty$, then $F(e) = 0$ and $E(e)$ is undetermined from the effort equation (see rows 3 and 6 in Table 1); however, unless $R(e_1) = R(e_2) = 0$, $E(e) = E(e_1 \,\vdots\, e_2)$.

For edges in parallel, $E(e_1) = E(e_2) = E(e_1 \| e_2)$ and the flow is $F(e) = E(e)/R(e)$. When $R = 0$ it is physically impossible that $E \neq 0$, unless there is a short circuit at the supply, which is treated as a special situation. If $R = \infty$ then $F = 0$ as in Table 1.

For $\mathsf{Y}$-$\Delta$ edges we have a sum of flows,

$$F_{ek} = \sum_{m=1}^{k-1} F(e_{mk}) - \sum_{m=k+1}^{n} F(e_{km}) \tag{8}$$

The addition of mixed signs leads to the possibility of an ambiguous flow (Table 3), as would be caused by a balanced bridge configuration. This indicates that the qualitative behaviour, and possibly, future state of the system depends on the numerical values of the resistances within one order of magnitude, signalling the analysis to try and obtain more detailed information. At the level of resistances, an ambiguous flow value does not necessarily lead to a reasoning impasse at the higher-level tasks such as FMEA if a higher-level behaviour is not dependent on the value and the analysis tool reporting does not require it. Finally, $E(e_{ek}) = F(e_{ek})R(e_{ek})$.

Figure 3 exemplifies a number of the concepts in the preceding subsections. The notation $\oplus$ and $\odot$ is used to identify positive and negative supply nodes, using a subscript to identify the source when necessary for systems with multiple power sources. Working from left to right in the Figure, a sequence of circuit reductions is performed to obtain a final single equivalent resistance value. An overall flow value is then computed and distributed amongst the circuit elements from right to left in the Figure as described above. Finally the flows in the $\mathsf{Y}$ are calculated from the $\Delta$ flows using equation 8, noting that the positive flow directions are defined as away from the star centre node.

$$
\begin{aligned}
F(e_1) &= -F(e_{14}) - F(e_{13}) = -f^{\triangleright 1} - f^{\triangleright 4} = -f^{\triangleright 1} \\
F(e_4) &= -F(e_{14}) - F(e_{43}) = f^{\triangleright 1} - f^{\triangleright 3} = f^{\triangleright 1} \\
F(e_3) &= -F(e_{13}) - F(e_{43}) = f^{\triangleright 4} + f^{\triangleright 3} = f^{\triangleright 3}
\end{aligned}
$$

## 2.4 Distinguished Node

Effort values are measured between two nodes and for convenience it is common practice to identify one distinguished node in a network and make measurements relative to that node. This then allows effort to be measured 'at' a node with the implicit assumption that the second node is the distinguished node. This is commonly the case in electrical systems where the distinguished node is called ground or earth and provides a reference node for voltage measurement and is often defined to be the negative supply terminal in a single source system.

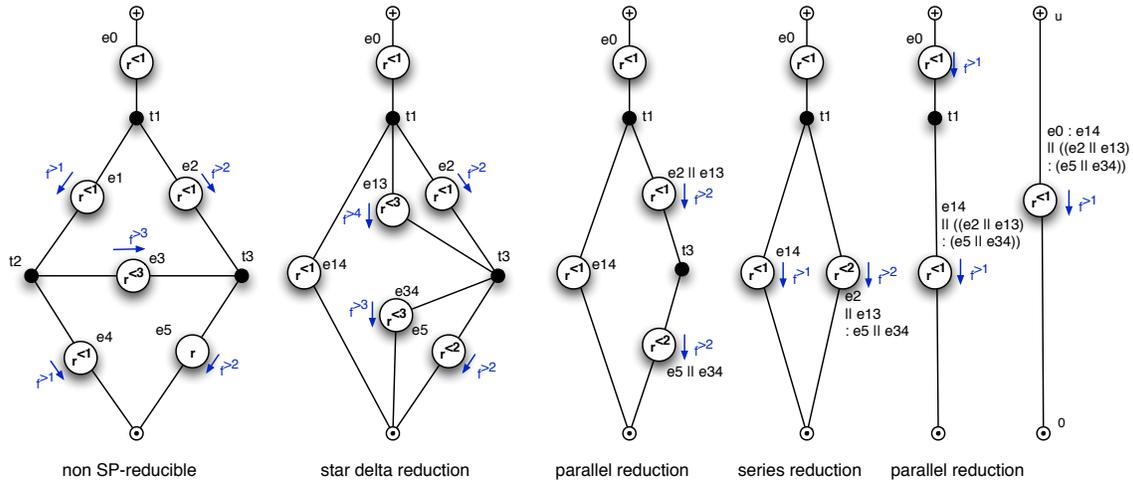

non SP-reducible        star delta reduction        parallel reduction        series reduction        parallel reduction

Figure 3: Operation of the qualitative circuit solver showing reduction and flow assignment

For generality we define the symbol $\mathcal{Z}$ as the distinguished zero node for a network of any type with specific instances, such as $\mathcal{A}$ to represent 'atmosphere' for fluid flow systems. Notice that $\mathcal{Z}$ does *not* provide another landmark value in the qualitative effort space; it provides a structural reference point in the resistive network.

The identification of $\mathcal{Z}$ allows the definition of 'absolute' effort relative to this point. The qualitative effort levels $[0, u]$ are supplemented – as for voltage (Lee, 1999) – with two additional symbols that represent structural features of the circuit that emerge during simulation. These additions are useful for interpretation of the simulation, but do not change the quantity space itself. $\emptyset$ is used to indicate a floating effort present in circuit fragments disconnected from any source and $\sim$ is used to indicate an effort that is between the supply terminal's voltage, i.e. $E(\oplus, \sim) = E(\sim, \odot)$ within the same magnitude. $\sim$ is associated to each source and is less useful for systems with multiple active effort sources. It is always useful to distinguish $\emptyset$ from 0 and *?* because $\emptyset$ is generally undefined and would not provide a meaningful measurement, although it *may* read 0 if measured. The qualitative value *?* is different from $\emptyset$ because *?* is qualitatively undetermined but has a value within the scope of the modelled system, whereas $\emptyset$ has no value within the modelled system.

## 2.5 Multiple Effort Sources

The network solver calculates power consumed ($P = E \times F$) for a single source. For a linear resistive network we can use a qualitative version of the principle of superposition where more than one source is connected to a single network (noting a single system schematic may, at any instant, actually comprise many isolated networks, possibly in different domains, even though all components on the schematic appear connected). Each network is analysed separately for each source by inhibiting all other effort sources and summing the results.

For sources $s_1 \ldots s_n$, the qualitative flow through an edge $e$ is given by $\sum_{s_1}^{s_n} F(e) = \max(F_{s_1}(e) \ldots F_{s_n}(e))$. Clearly where two opposing flows of the same magnitude exist, the edge suffers from qualitative ambiguity. For failure configurations, the ambiguous result is useful in highlighting to an engineer that a range of possible failure behaviour can occur. Simple relational constraints may be used to resolve an ambiguity for the common special case when there is a zero resistance between one of the supply nodes $\oplus_{s1}, \odot_{s1}$ and one of $\oplus_{s2}, \odot_{s2}$. For example a statement of the relative power of two pumps will allow the direction of flow in the reduced circuit to be derived, thereby allowing flows in the expanded network resistances to be calculated.

Figure 4 illustrates a circuit schematic for an electrical system with two effort sources and resistances of magnitude $r^{\triangleright 1}$ and r. The individual flow contributions calculated for source s1 are shown with a coarse broken line, and the flows for s2 with a fine broken line and are annotated with the flow magnitudes. The flow sum is shown for each resistor. There is partial flow ambiguity in the circuit, shown by the double-headed arrows next to resistors in the centre section where opposing flow contributions occur at the same magnitude. If the state-based or functional simulation requires the flow (direction), power or effort at these resistors, then it is necessary to know the values of the resistors.

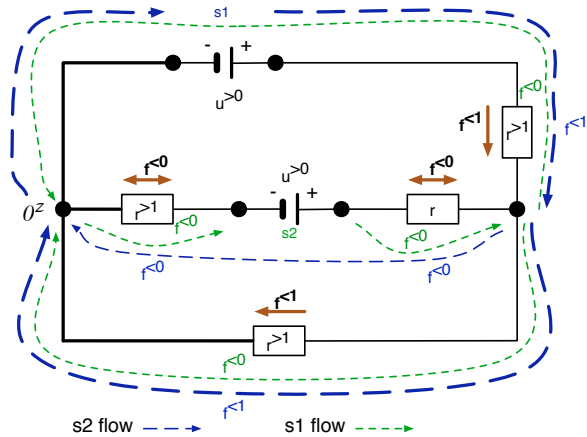

Figure 4: Multiple Source Example

## 2.6 Representation of Substance Within the System

The discussion so far has not considered the nature of the flow. For some domains the flow, such as electrical current, is implicit in the component models. In the thermal domain $E$ is a temperature difference and entropy flow is $F$. The product of these is 'heat' or thermal power, $P$, and the resistance of each component is selected to represent the reciprocal of thermal conductivity. Fluid transfer and hydraulic systems include the possibility that more than one substance can be associated with a network, particularly when faults are present. A concrete example is given by a fuel distribution system. If the supply tank becomes empty, air will enter the system and the behaviour of pumps and other component may change.

The substance-dependent behaviour is represented at the component level. The power network provides an instantaneous view of effort and flow, so it cannot participate in propa-gation of flowing substance, it is, however, necessary for the components to be able to obtain knowledge of the substance at the interface with other components. The network nodes are considered as zero volume points of connection that instantly propagate substances from component outflows to inflows. The local component behaviour provides substance information to the nodes in an appropriate qualitative timeframe according to the substance present at inflows, capacitance and other behaviour. There can be several connections to a node and flow direction may change during simulation. For these reasons a new concept is included in the circuit solver; a list of substances is maintained at each node that contains the output substances of each resistance connected to it at any instant of the simulation. Figure 5 depicts a node connected to three resistances. The flow directions are likely to cause *e1* and *e2* to modify the associated *SUBSTANCE* lists in *t1*. If *e3* then requests the substance from *t1* using $S(t1)$, then $\{S0, S1, S3\}$ will be obtained.

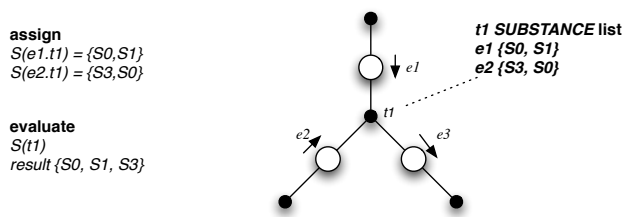

Figure 5: Node substance representation

The presence of more than one substance at a node will result in substance 'mixing' by virtue that all substances present are provided to any component that uses the node as an input. We might consider using the flow magnitude information to indicate the ratio of substance present, however, the reality will usually be that many other factors are involved, leading to a process of diminishing returns for the modelling effort required.

The modelling of substance in this work is deliberately simple and provides qualitative capabilities to match conventional bond graphs which address systems with free energy (mechanical, electrical, magnetic, incompressible fluid) in which all elements are conservative except the resistance, where energy is dissipated. Sophisticated techniques have been added to the Bond graph methodology (Brown, 2010), which allow numerical models of thermofluidic phenomena to be constructed; however, the complexity required of these models is such that the benefits of the broad-based qualitative modelling for system engineering analysis such as FMEA is lost. Some alternative approaches to qualitative reasoning have considered changes of material state for example a 'plug-based' ontology (Skorstad, 1992b) which is able to model the phase transitions in a steam boiler tube, however, the modelling required is necessarily detailed and the 13 state envisionment produced illustrates the complexity of the phenomena. Another approach to avoid complexity is used by Ghiaus who provides a qualitative model for the Carnot refrigeration cycle (Ghiaus, 1999) based on equations derived from a thermal bond graph and thus does not include substance properties as the system is assumed to be in equilibrium, precluding analysis of many failure modes.

Within our proposed ontology, compressible fluids require representation of the state of the substance and would require that the substance state is parameterised in the models.

Enthalpy flow could potentially be modelled by providing multiple forms of a substance (including phase changes) which are related to changes in containing volumes and pressures, providing effort to a thermal circuit. In any modelling enterprise, a choice has to be made as to the range of phenomena that are worth modelling, based on the required analysis. Complex thermodynamic aspects are one area where we have not considered automated analysis – at the overall system engineering level – to be worth the modelling effort because it is not clear that there is sufficient range of faults supported by models that can provide qualitatively distinct useful behaviour predictions. This is possibly an area for future research and may indeed provide fruitful results, but we make no claims at the current time.

### 2.7 Computational Enhancements

The network reduction and flow assignment detailed in the previous sections is sufficient to solve the OM effort and flow parameters for any network topology. There are, however, several additional concepts (Lee, 1999; Lee et al., 2001) that remain applicable using the SPS technique, and can provide additional vividness in the representation as well as computational benefit. These enhancements deal with very common special cases to avoid the need to use the Y-Δ transformation, which produces resistances that are not associated with the original circuit components in a straightforward way.

Any nodes connected by 0 valued resistances can be aggregated as a single supernode, with $E(e) = 0$ across any edges subsumed by the supernode. For an edge $e = \langle t_i, t_j \rangle$, where $t_i, t_j \in T$, and $R(e) = 0$ a supernode $t_{ij}$ is created with the label $t_i.t_j$, and the edge is removed from the active graph. Once generated a supernode can participate in further SP Star (SPS) reduction, however, upon expansion of the circuit it is not possible to directly allocate flow to the edges represented by the subsumed edges as is done with SPS expansions.

The flow within supernode edges can be deduced for unambiguous (non-ladder network) cases using a qualitative version of the Kirchoff's current law. With the exception of the source nodes, for a node $t$ and connected edges with flow $F(e)$:

$$\sum_{\{e \in A : e = \langle t, x \rangle \ \lor \ e = \langle x, t \rangle\}} F(e) = 0$$

For supernode edge flow values, $f_1 {}^{\rhd m_1} ... f_n {}^{\rhd m_n}$, flows $F' = \{f_x {}^{\rhd m_x} : m_x = max(m_1...m_2)\}$ dominate. For a node with $|F| - 1$ edges flowing into the node the unassigned flow edge must be out of the node with a flow equal to the flow magnitude in $F'$ towards the node. The dual exists for flows in the other direction.

Two additional enhancements may be used to improve implementation performance of the network analyser. Edges $e = \langle t_i, t_i \rangle, t_i \in T$ are a loop and can be removed from the active graph since the loop is by-passed by a zero resistance path, hence is assigned zero flow upon expansion. Loops typically represent shorted out parts of an electrical circuit for example.

Degree one nodes, $e \in \{\langle t, \emptyset \rangle, \langle \emptyset, t \rangle\}, t \in T$, are known as dangling edges and are also removed from the graph together with the connected edge as and when they occur as a result of other reductions.

## 3. Local Component Models and OM Time

The resistive network calculates power consumption, $P$, but cannot model any component that stores energy, $En$, since $En = P \times T$ for time $T$. Displacement and momentum are the two domain independent characteristics resulting from the inclusion of time into the model (Figure 1). Energy, displacement and momentum are fundamentally local characteristics of components derived from the power variables and time. The representation of time follows the OM approach described in Section 2.1. For a time $t^{\rhd n}$ and flow $f^{\rhd n}$, $d^{\rhd (n+m)} = f^{\rhd n} t^{\rhd n}$ defines a displacement $d$ such as quantity of substance.

A FSM representation of local component behaviour has proved sufficient for abstract behaviours for the failure analysis task. Time is represented by state changes and is explicitly represented by state transitions that capture the qualitative integration of effort or flow variables. The state of the power network is used to trigger transitions, as are input events from outside the system (external interactions). The change of state of a component may cause the structure or resistive parameters of the power network to change, thus triggering a power network simulation and a sequence of events in other components. This results in a system of interacting state machines, sharing time as a common variable that sequences events.

We use a specialised subset of the UML language state chart notation (OMG, 2012) to describe the FSM models. The model is comprised of a set of states $S$, events $\Sigma$, and transitions $\delta = S \times \Sigma \to S$. In addition $s_0 \in S$ defines the initial (default) state of the component. Output actions $A$ may be associated both with $e \in \Sigma$ and also as entry actions associated with $s \in S$. The UML provides for guard conditions on events and we refine these conditions to produce $e = (t, T_c, D_c, F_c, A)$. $t$ represents a temporal condition for the transition as a qualitative OM duration after which the transition can occur once $T_c$ is satisfied, provided $F_c$ is true after this time. i.e. $T_c$ triggers a transition and $F_c$ and allows it to complete (fire). $D_c$ is a condition that must be satisfied during the transition, between the satisfaction of $T_c$ and $F_c$. All conditions may appeal to values from the network model, and $A$ may change the resistance values of the network model. States are used to represent the qualitatively significant values of variables that are derived from the integration of efforts and flows over time or higher-level states of components. For example capacitor charged/discharged, or relay activated/deactivated. The following sections describe the component level simulation and the component model syntax.

### 3.1 Component Level Simulation

The presence of multiple components in a system results in a set of independent interacting FSMs. The only global variable at this level is time, and the OM representation provides for sequencing of component behaviour at different timescales. The definition of OM applied to time requires that a sequence of (non-cyclic) events in $t^{\rhd (n-1)}$ occur before an event in $t^{\rhd n}$.

The processing of events is carried out by maintaining a time-ordered priority list of all component events with satisfied conditions. All events are ranked by their order of magnitude time delay periods (referred to as *time-slots* subsequently) in the following way:

- All events $e$ where $T_c$ is satisfied are added to the end of $Q$ for the priority specified by $t$ in the event or 0 if unspecified.

- Candidate events to fire are $n \in N$ where $N \subseteq Q$ and $t(n) = t^{\rhd x}$ such that $\min(x) \land N \neq \emptyset$, i.e all the events in the lowest order non-empty timeslot.

- All events $n$ where $F_c(n)$ is satisfied are fired, and events where $\neg F_c(n)$ are removed from the queue.

- Any events where $D_c$ if not satisfied are removed from the queue.

When $|n| = 0$ the system has reached a steady state and there are no further changes of state. When $|n| > 1$ there is non-determinism in the system. The important question for FMEA is the longer term impact of the alternative behaviours on the potential worst case faults and this usually depends on whether the alternative behaviours diverge to significantly different functional (external) effects or are alternate paths through internal states that converge on a common state. Very often it is the latter case. For example two relays wired in parallel may not switch at exactly the same moment, resulting in two possible behaviour paths and two intermediate states, neither of which is significant in most cases.

In some systems it is possible to make a concurrency assumption that specifies that the final state reached at the end of time period $t^{\rhd x}$ is independent of the ordering of the events in the time period. The presence of race conditions or feedback loops between mutually interacting components does not allow the concurrency assumption, allowing the system to reach a qualitatively distinct state dependent on the detailed numerical timing of the events. This is a case where the qualitative representation of time lacks enough detail, and this qualitative ambiguity will not be detected if the concurrency assumption is falsely applied. In most systems it is certainly reasonable to assume that all $t = 0$ events are concurrent, provided only events associated with the power network have $t = 0$ because these rarely have causal cycles, unless specific examples such as bistable logic gate configurations are created. Usually such 'memory' features would be represented at a higher-level as the state variables (e.g. electronic control unit), leaving the domain-based modelling with non cyclic causality.

Generally, each possible ordering of the events in each time-slot must be considered to determine if all branches (eventually) reach the same state and, if this is not the case, generate a number of alternative behaviour paths. This is achieved by a breadth-first search to determine converging or cyclic behaviours when the system reaches a state that is *identical* to a previously encountered state. There is no assumption made about such behaviour; it is necessary for the simulation to detect cyclic behaviour to allow termination and appropriate reporting. Alternatively the simulation must seek additional information or advice from the engineer, if, for example, no converging state is found within a reasonable length behaviour path.

## 3.2 Component Modelling Examples

A graphical notation is used to describe component models, for example Figure 6. State entry actions are placed inside the rectangle representing the state, and an event syntax '[if $T_c$ [during]] event_name [after $t$ [$F_c$]][/$A$]' is used where [$x$] represents an optional element $x$. The keyword during is used to specify that $D_c = T_c$, and also that $F_c = T_c$ if $F_c$ is unspecified. If during is omitted then $D_c = \top$. If $F_c$ is omitted then $F_c = \top$. The keyword after is used to specify $t$. If no after is specified, then the event is at $t = 0$, i.e. immediate occurrence.

As an example assume we have the following OM time values $F = \{\mathsf{mS}, \mathsf{Sec}, \mathsf{hour}, \mathsf{day}\}$ and flow levels $T = \{\mathsf{low}, \mathsf{normal}, \mathsf{high}\}$, where Sec and normal are given the OM index of zero. A tank of a given volume may be defined by an event that changes from the 'empty' to 'full' state in a given time, e.g. 'if $F(\mathsf{tank\_inlet}) == \mathsf{normal}$ during filling after hour' provides an implicit volume of $\triangleleft 1$, or one order of magnitude bigger than a nominal volume, given the chosen flow and time qualitative space. By explicitly including the volume in the event conditions it can be made to represent a number of possible transitions after different durations, for example, 'if $F(\mathsf{tank\_inlet}) > 0$ during filling after $\mathsf{tank\_volume}/F(\mathsf{tank\_inlet})$', where $\mathsf{tank\_volume}$ is defined by the component to have the qualitative volume value with magnitude $\triangleleft 1$. The flow condition prevents an event that requires an infinite amount of time to fill the tank ($0$ is effectively $\mathsf{normal}^{\triangleleft\infty}$)

Figure 6 shows the two-part model for a tank. The two levels of the model are shown on the left, with graphic icons on the right that can be used to display simulation results on any schematic containing an instance of this component. The structure is a single zero resistance because the tank dissipates no energy during filling. The capacity of the tank is defined by a local component variable volume, used in combination with the flow, $\mathsf{F(tk)}$, to control the change of state, allowing several different capacity instances of the tank to be created by parameterising the model.

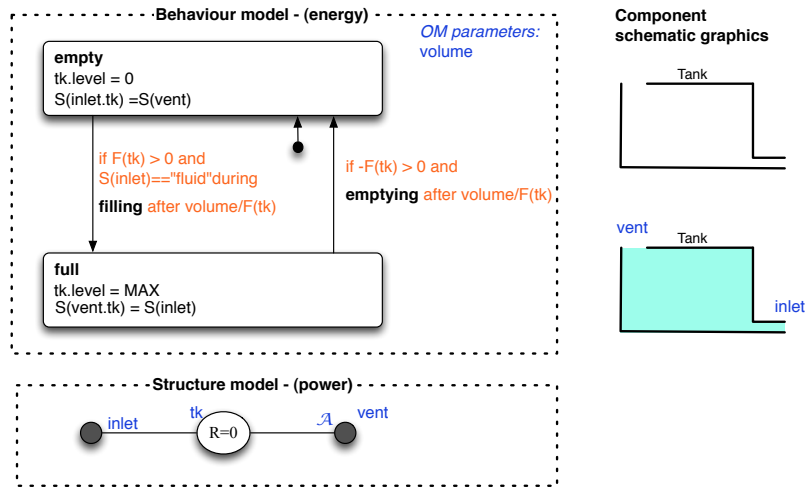

Figure 6: Basic 3 tank, with no energy storage

The tank in Figure 6 does not store any energy because the effects of gravity are ignored and hence there is no increase in potential energy as the tank fills. This aspect could be included in the modelling by making the tank into a small magnitude effort source when its level is non-zero. It will then include capacitance as well as displacement (zero stored energy volume) provided by the duration the flow continues prior to change of state. Atmospheric pressure is often a significant qualitative value in fluid flow systems and the behaviour of a system may depend upon the pressure difference between atmospheric pressure and some other point. $\mathcal{A}$ is defined as the fluid flow domain specialisation of $\mathcal{Z}$ and provides a global node accessible from any component model that represents a connection to the atmosphere. Examples include a vented tank or leaking pipe.

A different component is used to illustrate a component containing a dependent effort source. Figure 7 shows a model for a (non-horizontal) pipe that contains either liquid or a gas (air). A resistance of $R(\texttt{pipe})$ limits the flow in the pipe and represents a combined measure of smoothness, length, Reynolds number etc. The pipe model is parameterised; an ideal pipe is produced if $\texttt{length} = 0$. A pipe blockage fault model could set $R(\texttt{pipe}) = \infty$.

The filled pipe represents a small magnitude effort source. The pipe has an additional resistance to represent the 'pipe wall' to facilitate open fault modelling by providing a possible connection to $\mathcal{A}$. The event conditions cause the pipe to fill at a rate inversely proportional to the flow through it. The rightmost event specifies that the pipe can prime itself, ie. if fluid is present at the top inlet (only), it will fill without externally imposed flow.

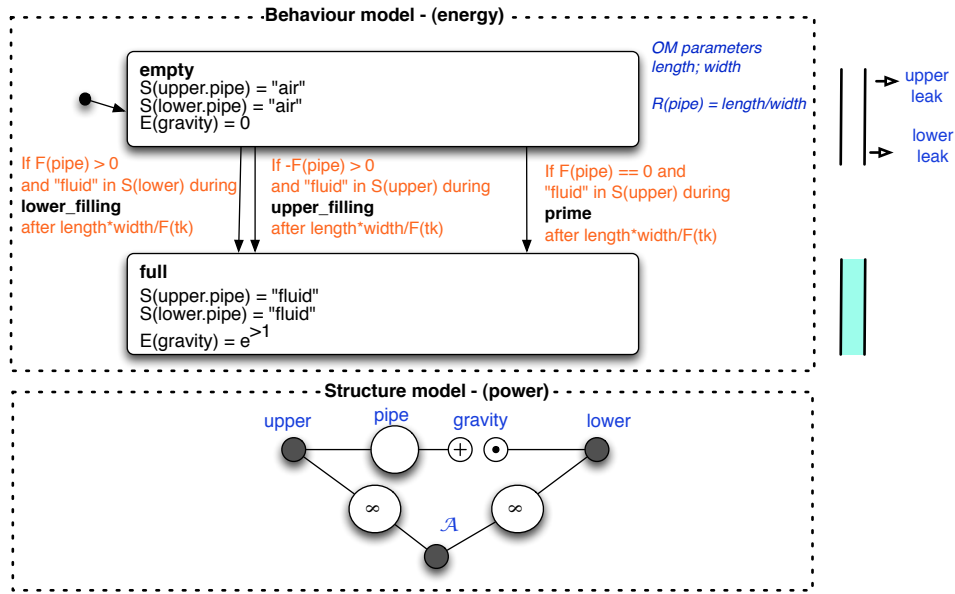

Figure 7: Partial model for an vertical pipe

Figure 8 illustrates some general ways a component local behaviour may interact with the structure for several common component types. Real component models will have additional relationships and constraints associated with qualitative values and device states, dependent on the details of the behaviour required.

## 3.3 System Modelling and Circuit Topology

Figure 9 is an artificial system of components similar to those described in the previous section and will be used to illustrate topological aspects of the modelling and simulation. In addition a valve with closed ($R = \infty$) and open ($R = 0$) states related to an external position input is included, together with a simple pump that may be activated via 'activate' and 'deactivate' external events. If the electrical aspects of the system were included then the pump model would include an electrical resistance and the events would be triggered by the level of power (or current flow) in the electrical resistance, thus setting the effort level of the pump appropriately.

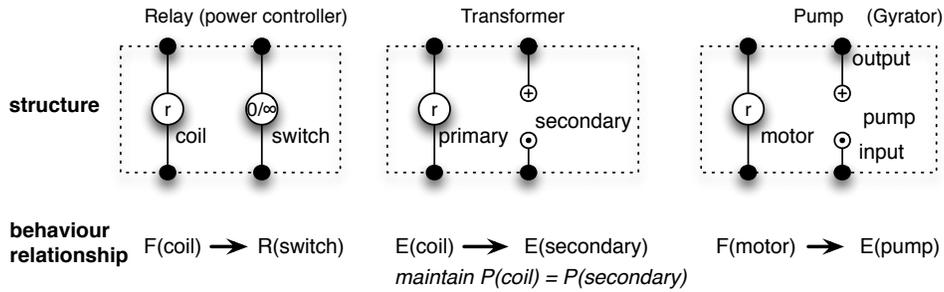

Figure 8: Local/global relationships for several types of component

Other qualitative aspects of the pump can easily be included in the model to create different types of pump. The details of the design of the pump may allow flow when it is inactive as is the case here, however, it is easy to include a state-controlled resistance set to $\infty$ for a pump that does not allow flow when inactive. Similarly, with an additional condition on the events, the pump can be made non-self priming if no fluid is present at the input. A bi-directional pump type requires an additional state and selection of the correct substance input.

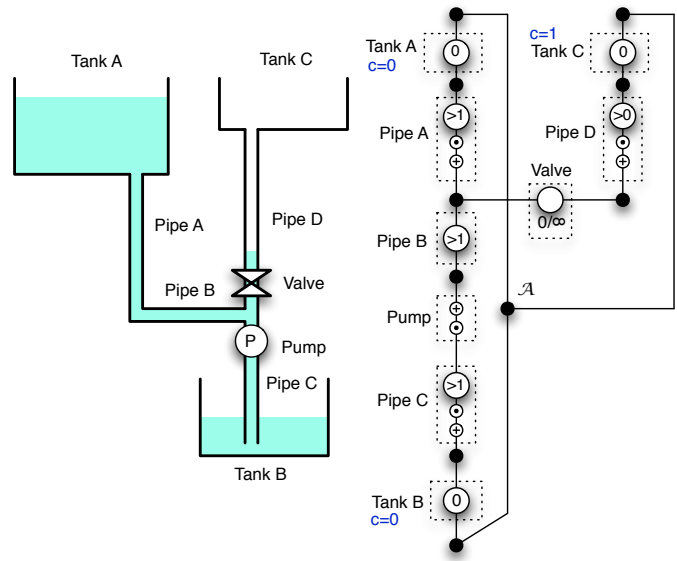

Figure 9: Pumped System

Consider the simulation of the system in Figure 9 from the state at the top of Table 4 with the pump off. The table summarises the changes of state of each component with horizontal lines at the points where the network simulation takes place. The Pump Valve Pipe D, do not change state maintaining the states off, closed and empty respectively. In this example there are no resistance changes, only changes to effort sources. Initially the vertical pipes prime and fill in sequence as fluid is propagated from inflow nodes to outflow nodes, creating two pressure sources that cause a flow from the upper to lower tank and a flow of air out of the vent of the lower tank and into the upper tank.

| Tank A | Tank B | Tank C | Pipe A | Pipe B | Pipe C |
|---|---|---|---|---|---|
| full | empty | empty | empty | empty | empty |
| | | | →prime, $t^{\triangleright 2}$ <br> full, $\oplus = u^{\triangleright 1}$ | | |
| →start_emptying <br> part_filled <br> [→emptying, $t^{\triangleright 1}$] | | | | →prime, $t^{\triangleright 2}$ <br> full | |
| | | | | | →prime, $t^{\triangleright 2}$ <br> full, $\oplus = u^{\triangleright 1}$ |
| | →start_filling <br> part_filled <br> [→filling, $t^{\triangleright 1}$] | | | | |
| Chronological non-determinism in time $t^{\triangleright 1}$: ① Tank A →emptying; ② Tank B →filling | | | | | |
| **Choose > ①** <br> [→emptying, $t^{\triangleright 1}$] <br> empty | | | | | |
| | | | →drain, $t^{\triangleright 2}$ <br> empty, $\oplus = 0$ | | |
| | | | | →drain, $t^{\triangleright 2}$ | →drain, $t^{\triangleright 2}$ <br> empty, $\oplus = 0$ |
| **Choose > ②** | **delete** [→filling, $t^{\triangleright 1}$] <br> ¬($S$(inlet)=="fluid") | | | | |
| | [→filling, $t^{\triangleright 1}$] <br> full | | | | |
| **Fluid present at** $\mathcal{A}$ <br> [→emptying, $t^{\triangleright 1}$] <br> *continues as for* ① | | | | | |

Table 4: Simulation sequence for system in Figure 9

A qualitative ambiguity then results between two events of the same duration, raising a question which could not be answered without knowing the numerical sizes of the tanks (and the quantity in each if not full and empty at the start). In this case the simulation was allowed to try the alternative behaviours. The first is that Tank A becomes empty before Tank B fills and a steady state is reached with Tank A empty and Tank B is part filled. The second possibility is that the lower tank becomes filled before the upper one empties. This causes fluid to reach the $\mathcal{A}$ node, and this is built into a model of the atmosphere that reports an abnormal condition because a substance other than 'air' is present. Finally a steady state is reached where Tank A is empty and Tank B is full.

If the pump is switched on and the valve opened, the pump effort source causes flow up to both top tanks because the pump is a larger magnitude of effort than gravity in the pipes, however Tank C is an order of size larger and the simulation is essentially the reverse of previous example. Tank A overfilling is a spurious prediction because the simulation is not able to reason that all of the fluid originally came from Tank A.

An illustration of flow ambiguity occurs if only Tank A is full and the valve is open and pump off. The simulation starts as before until Pipe B is filled, then Pipe D becomes filled from the bottom due to the effort from Pipe A. Pipe D then becomes a pressure source and is ambiguous with the opposing effort from Pipe A. The $\oplus_{\text{PipeA}}$ and $\oplus_{\text{PipeB}}$ supply nodes are both connected to a supernode allowing specification that $E(\text{Pipe D.gravity}) < E(\text{Pipe A.gravity})$ to establish flow direction; Pipe D will fill and there will be a flow into Tank C, as would occur if Tank C was lower than Tank A. It is large so never becomes full, and subsequently drains through PipeD.

If $E(\text{Pipe D.gravity}) > E(\text{Pipe A.gravity})$, then Pipe D drains itself (producing air at the valve) whereupon the cycle repeats. The simulation concludes that the pipe is oscillating between the full and the empty state and its state is undetermined at this modelling resolution. There may be physical oscillation, however, usually this indicates that the system is in a state that cannot be represented at the level of abstraction being used.

The final possibility is that $E(\text{Pipe D.gravity}) = E(\text{Pipe A.gravity})$, (it has filled to the same height as Tank A) then there is no flow in Pipe D. In all cases there is flow to Tank C which is either full or part filled at steady state.

These last examples were deliberately chosen to be at the limit of the representation, if they occur during an FMEA during nominal operation, there is a clear indication that a more detailed numerical model is required for the state or behaviour; if they occur during a component fault then there is a signal that a failure mode has been encountered that requires further detailed investigation.

An equation-based qualitative model of a system of tanks (Dressler, Böttcher, Montag, & Brinkop, 1993) allows some comparisons to be drawn with our approach. Firstly, the authors note an important feature also relevant to our approach; for diagnosis (and also FMEA) a complete behaviour description is not necessary. In fact, a model containing too much detail is likely to produce many qualitative ambiguities (or branching behaviour) on unimportant behaviour aspects. These problems have been described in detail (Clancy, Brajnik, & Kay, 1997) and strategies to avoid the problem such as *model revision* proposed, with various tools used to support the revision of (QSIM) models. While this may eventually result is a desired simulation result, such strategies prove problematic for FMEA due to the range of behaviour likely to be encountered by the automatic insertion of component failure models into the system; we need abstracted models that are nevertheless grounded in the basic physics and are fully compositional.

To avoid such difficulties a simplified numerical model containing the relevant behavioural phenomena, can be used (Dressler et al., 1993) and subsequently refined this into a qualitative one. During this refinement process, variables and values were mapped onto qualitative values $(0, +, \infty)$; however, some values were required to be treated differently with various landmarks being identified –namely the height of liquid in the tanks and the atmospheric pressure. This illustrates the problem using the general equations, since a numerical equation does not provide landmarks that represent states such as the volume of a tank, and global features such as $\mathcal{A}$. We suggest that explicit state identification, with temporal state changes, provides a more vivid qualitative model, within a broadly applicable domain-independent generalised framework. Other components used by Dressler, such as a valve, require mappings between an input command and the flow in the valve, using an expression to set the valve flows (denoted i) at input and output as follows: valve.status = :close → valve.$i_1$ = valve.$i_2$ = 0. We also note from the model that an open valve is defined as valve.$i_1$ = −valve.$i_2$ and therefore propagates the flow locally when the valve is open. This requires a predetermination of cause and effect in the power network, unlike our global network model. In our approach a state description of the valve (relay in Figure 8 is analogous to an electric valve) that controls the qualitative resistance is more vivid (and physically correct) than controlling the flow, since the flow depends on pressure, and pressure cannot be determined locally. If an OM model is used, a partly blocked valve can be represented by an OM higher resistance than normal. In this situation it is not feasible to set the flow value based on valve control state, because it depends on the external system.

The $\mathcal{A}$ node is important to the circuit-based representation in the pumped system example, however in general the $\mathcal{Z}$ node may also lead to power network ambiguity unless the following condition is satisfied. The zero node must partition the graph into two disjoint sets of edges sharing only $\mathcal{Z}$ and supply nodes. This topology naturally exists in many practical fluid flow circuits, because the only connection between the negative (suction) and positive (pressure) parts of the system are the pump itself and the atmosphere. For others such as the contrived example in Figure 10, analysis is inherently limited since it is impossible to determine qualitatively that a leak in Pipe E would cause liquid egress or air ingress. The ambiguity is indicated because the direction of flow through the leak (represented by the dashed resistances on the right of the figure) at either end of the pipe is different with respect to the atmosphere, as shown by the arrows.

The example also illustrates two additional resistances used to represent a 'leak' failure mode because of the qualitative difference in behaviour dependent upon leak position. The resistance magnitude is used to indicate the severity of the leak where zero would produce a complete fracture. The arrows on the leaks from PipeC and PipeD demonstrate an unambiguous qualitative behaviour, however PipeE shows conflicting leak flows with respect to $\mathcal{A}$. This qualitative result is exactly what we would expect in the absence of numerical information, and explicitly signals the limit of the behaviour predictions possible with limited information about the system. Circuits that have resistances that bridge the $\mathcal{Z}$ node also cannot have effort values assigned relative to $\mathcal{Z}$.

## 4. Faults and Exaggeration Reasoning

Exaggeration reasoning (Weld, 1988b, 1990, 1988a) provides an alternative qualitative technique for explanation of the worst case effects without the need for differential qualitative calculus. This form of reasoning is therefore suitable when we do not have detailed system equations, or, as for the global power network, reasoning about the causality is not performed. A secondary advantage for our purpose is that exaggeration reasoning often produces more concise explanations. The drawback, however, is that qualitative deviation analysis is guaranteed sound, whereas exaggeration reasoning can lead to false predictions.

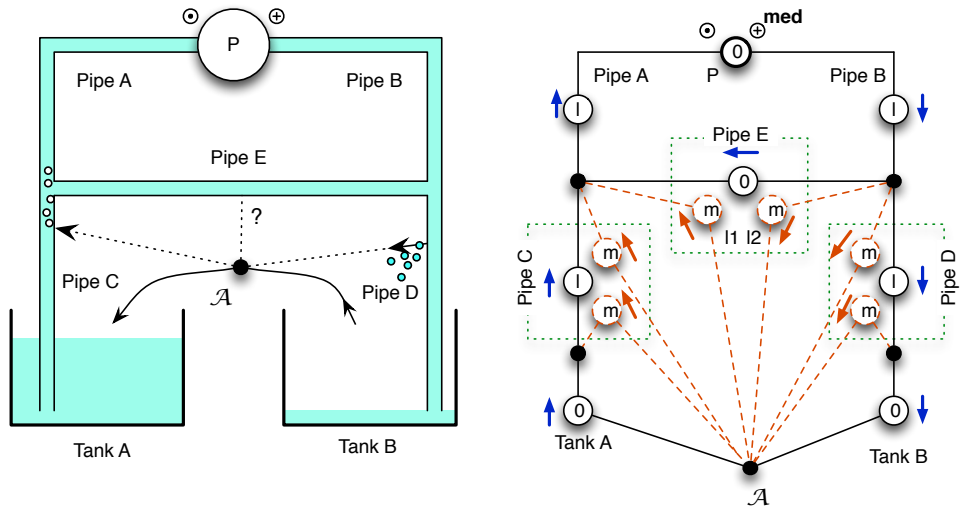

Figure 10: $\mathcal{A}$ not forming disjoint graphs

For the purpose of FMEA there are two reasons why exaggerating faults is a reasonable strategy: firstly, the worst case effects are required, which results in fault models at the extremes of behaviour; secondly, the FMEA is intended as an aid to the engineer and is therefore externally verified.

Using a simple electrical torch as an example, a corroded battery contact fault may raise the contact resistance and the torch may dim, however, this resistance increase does not change the qualitative behaviour. An exaggerated fault would be an OM increase in the battery contact and the simulation predicts an OM reduction in light output. Of course a OM reduction of light would likely not be visible, but for the FMEA the effect that corrosion leads to a reduction in light is reasonable and provides a significant distinction from a fracture effect where no light or circuit activity is present.

The OM representation provides for exaggerated forms of faults that model significant differences in behaviour. For example a small leak in a fluid system pipe may allow air to be sucked in, causing a mixture of fluid and air at the output; a fracture in the same position may result in no output because the pump fails to operate with only air.

A common approach to reasoning about this qualitatively is to use a qualitative constraint-based deviation model and propagate the deviation through the system. Reasoning about deviation works well for parameter value changes, but is less good where structure or state changes occur, such as the air ingress example above where a new system state occurs due to air in the pump. Each approach has its strengths, and so we might use absolute/exaggeration reasoning to determine the impact of faults on major states and operating modes (or regions of linear behaviour), followed by deviations to isolate finer grained effects such as expected direction of value drift due to a fault.

## 5. Generating an FMEA

An automated FMEA is typically based on behaviour simulation for many component faults and operational scenarios. Producing the FMEA has been previously described in detail elsewhere (Price, 1998), however, since it is the end goal of the modelling and simulation effort, we summarise the main steps as follows:

- The system is simulated with no failed components over the expected operating conditions.

- The system is simulated for each of the component failure modes contained in the component type models for every component instance, over the same operating conditions. Multiple faults may also be considered for high failure rate component combinations.

- The output of the simulation is the qualitative value of all system variables for each step (state) of the simulation. These are used by a completely separate functional model that identifies specific (output) behaviour with the state of identified system functions. Functions can be in one of four states (Achieved, Failed, Unexpected Behaviour, Inoperative) based on the truth of Boolean trigger and effect expressions that evaluate simulation output variables. The functional model is lightweight, but can capture a hierarchy of system functions, including temporal aspects. Full details of the functional model have been previously presented (Bell, Snooke, & Price, 2007).

- The nominal and failure functional states are compared, and used to indicate at a high level of abstraction, the highest risk failed system functions and unexpected function effects associated with each component fault.

- Further presentation, selection, ranking of the function states and risk factors computed from information associated with the function states and component reliability allows an FMEA to be produced. The structure of a typical automatically generated FMEA is shown in Figure 17 for our second example, discussed in Section 7.

It is the simulation step of this process that is of interest in this paper and the following sections provide two example systems as illustrations.

## 6. Case Study Example: Domestic Heating System

Figure 12 shows a schematic for part of a simple domestic central heating system. The complete model includes an electrical microcontroller that controls voltage to the pumps and activators, in addition to a thermal system, we however focus on the fluid flow aspect. The gas boiler and three way valve had been retrofitted some years after initial installation. Either the gas boiler or wood burning stove can supply hot water. The radiators are modelled as $R = r$, the pipes are ideal (`length`=0) and the boilers and hot water cylinder as $R = r^{\rhd 1}$ because they are much larger diameter than the radiators. The components also include a thermal element where flow is considered as the entropy flow rate and temperature as effort. The connection between these components is a compound connection including the fluid and thermal circuits. There is not space to detail the complete set of models, however, Figure 13 shows the horizontal pipe model that includes both the fluid and thermal aspects.

The fluid flow element is resistive and propagates the fluid based on flow direction. The thermal element is represented by a further resistance. The length of the pipe determines the thermal resistance of the pipe by conduction when there is no flow, and when there is flow, there is no thermal resistance because the heated substance is transported. Thermal power is provided by the boiler (by combustion) to create a temperature difference (effort) between inlet and outlet. This thermal effort is then applied across the radiators which provide power (heat) to the surrounding air.

A further aspect of the heating system is the interaction between the fluid flow and temperature networks. For a flow rate such that the returning temperature of the fluid is substantially higher than the ambient air, series connected radiators each consume a ratio of the thermal power based on their dimensions. For low flow rates this assumption may not hold. For example, to represent the situation that a low fluid flow will allow more thermal power to dissipate, an additional resistance is included in the thermal radiator and pipe models that allows *direct specification* based on the low flow rate, such that the outlet temperature is at or close to the inlet temperature of the boiler (or atmosphere). In Figure 13 this is the low_flow resistance, controlled by the flow. Figure 11 shows the effect of a low fluid flow rate, on a series of radiators. The low_flow elements have changed from $\infty$ to $r^{\triangleright 1}$. The power decreases by OM from the source. The first radiator is hot (if the effort source is higher because the flow is lower), the next one is an OM cooler, and so on.

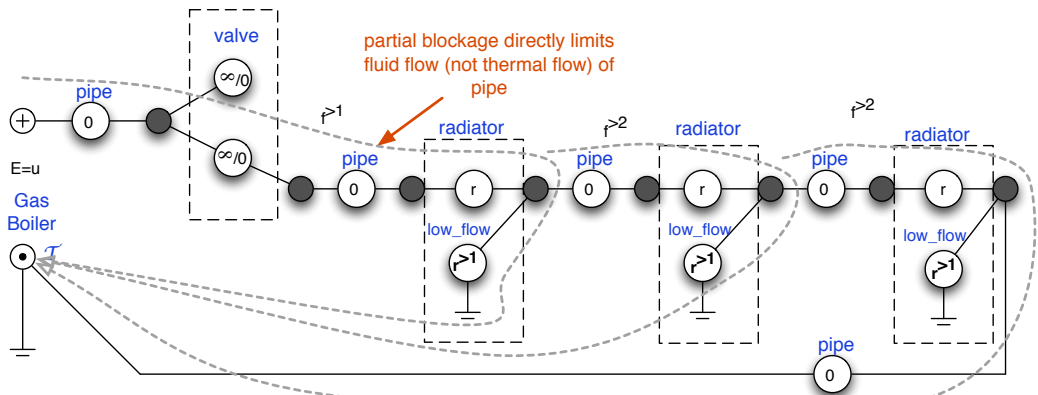

Figure 11: Thermal flow circuit in a low fluid flow situation

Consider a small leak in vpipe0 directly below the pump + output. In Figure 12 during operation of the gas boiler with the valve in the heating position, fluid flows through the radiators and heat is transferred to the radiators. A small amount of fluid enters the atmosphere, causing a small flow of fluid from the header tank which is replaced by water from the external water supply.

Now we try a holiday scenario with the simulation results summarised in Table 5. The external water supply is isolated (in case of freezing). The flow model with the pump off, derived from the component models and their connections, is in Figure 14. There is an ambiguity concerning the gravity sources in the pipes p1, p7, p11(names abbreviated) which oppose p0, p3, p6, p12, p13 at the same flow magnitude. The user can provide that the relationship between $E(\mathtt{vpipe3.gravity})$ and $E(\mathtt{vpipe11.gravity})$, is "=" and similarly for the other two sets ($\mathtt{p6}$=$\mathtt{p7}$, $\mathtt{p3}$=$\mathtt{p11}$, $\mathtt{p1}$=$\mathtt{p0}$+$\mathtt{p13}$+$\mathtt{p12}$). The result is a flow from the header tank to the leak via $\mathtt{p16}$, $\mathtt{p12}$ and $\mathtt{p13}$. Air then enters $\mathtt{p16}$ and it is no longer an effort source (Table 5, row 3). In addition a secondary low flow exists via $\mathtt{p1}$ and $\mathtt{p0}$, these flows are in opposition to each other, however the constraint above allows the $\mathtt{p1}$ source to predominate.

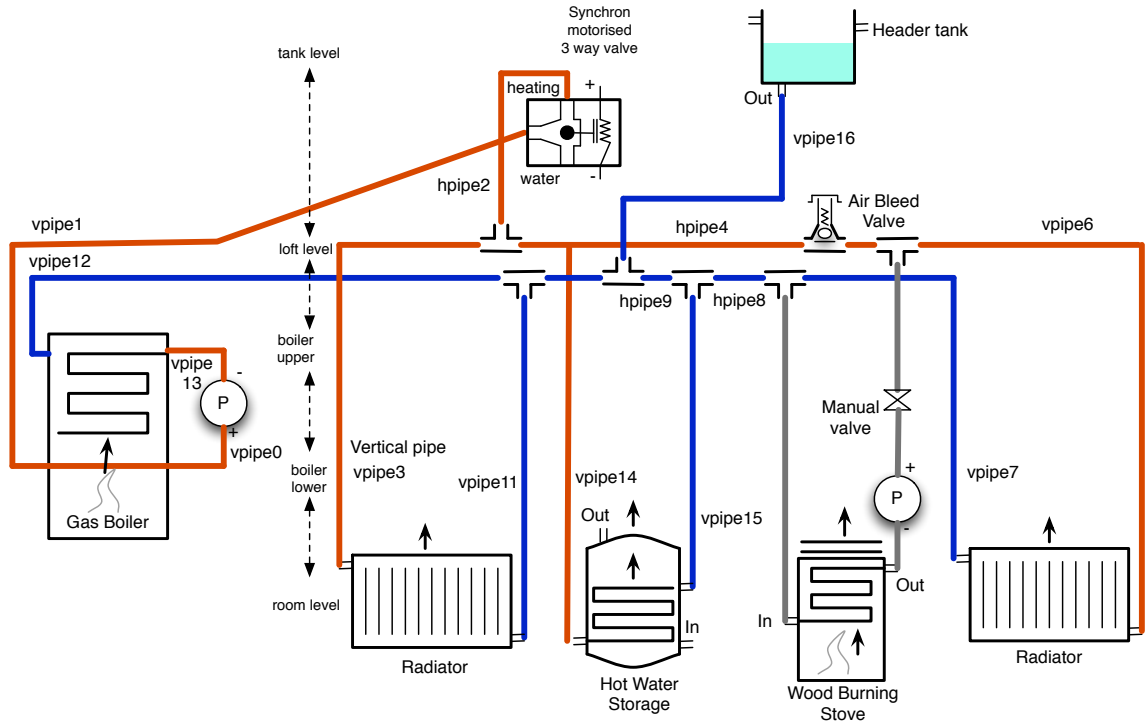

Figure 12: Domestic heating system

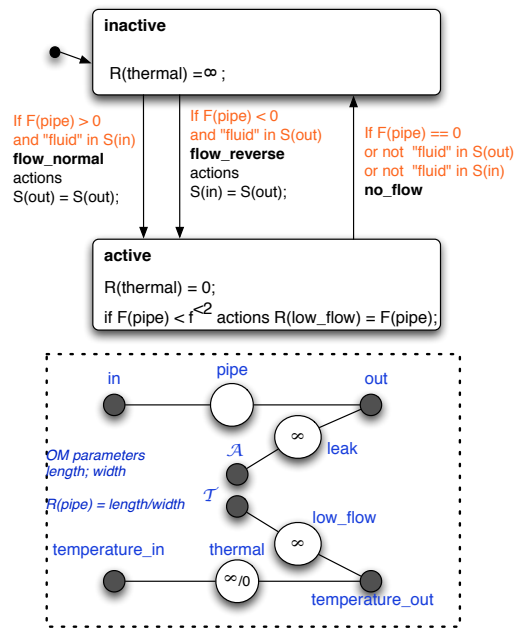

Figure 13: Horizontal pipe with thermal aspect

Now there are four sources ($\mathtt{p0}$, $\mathtt{p1}$, $\mathtt{p12}$, $\mathtt{p13}$) and it turns out there are three flow patterns. Figure 15 shows the flow system with all zero resistances and all dead branches removed, using a different style arrow to show the contributing flow patterns. The radiator section has higher resistance and therefore the main flow is due to effort from $\mathtt{p12}$ and $\mathtt{p13}$, drawing air from $\mathcal{A}$ through the boiler, $\mathtt{p13}$, and pump to the leak until air reaches $\mathtt{p12}$ and its effort becomes zero. The secondary flow pulls air toward the radiators until the cold side pipe ($\mathtt{p11}$) becomes empty, whereupon there is an ambiguity between effort from $\mathtt{p3}$ and $\mathtt{p0}$. By specifying that the gravity sources $\mathtt{p3}$>$\mathtt{p1}$, the flow direction reverses and $\mathtt{p0}$ and $\mathtt{p1}$ fill with air, $\mathtt{p11}$ fills with fluid from $\mathtt{p3}$. Since the pipes all have the same length and diameter there are several possible behaviours; in the table we provide additional information rather than generate the alternative behaviours – which would lead to uncertainty as to the fill state of $\mathtt{p0}$, $\mathtt{p1}$, $\mathtt{p2}$. Now $\mathtt{p0}$, $\mathtt{p1}$, $\mathtt{p2}$, $\mathtt{p12}$, $\mathtt{p13}$ contain air. $\mathtt{p3}$, $\mathtt{p11}$, $\mathtt{p6}$, $\mathtt{p7}$ contain fluid and are opposing equal sources and the system is stable.

Returning home and turning on the water causes the header tank to refill, however, the pump does not produce flow because it does not self prime. $\mathtt{p16}$ will prime and create a low flow from the header tank but will take $t^{\rhd 1}$ to propagate to the pump. Hopefully it is not winter.

In a subsequent step in the FMEA scenario the wood burning stove pump is started and valve opened. A large flow of water is pulled from the header tank through the three way valve the 'wrong way', air and water mix in the return pipe and are pumped past the air bleed valve, where the air is removed. The gas boiler and pump can be restarted and the heating system works correctly - other than a small flow from the header tank to the leak.

The resulting FMEA report will highlight the effect of the failure – no heat output at any radiator after the holiday scenario step and that this is not as severe as other faults (a burner fault for example) because it is not permanent.

## 7. Case Study Example: Aircraft Fuel System

This section describes a fuel system provided by the sponsor of the work. The modelling and simulation described forms part of a wider objective to automate manually created FMEA reports used to generate fault effect relationships as input to a Bayesian network based diagnostic system. This effort is, in turn, part of a recent 32m UK government sponsored programme seeking to research, develop and validate the necessary technologies for use in unmanned aerial systems (ASTRAEA, 2009).

The system considered is a fuel system for a twin engine light aircraft shown in Figure 16 and was also available as a physical laboratory simulation on a configurable rig that allowed validation of the results using fault insertion via additional components such as valves to represent leaks. The requirements for this system did not include any thermal aspects and due to the orientation changes of the system only pressure created by the pumps was required, resulting in very little qualitative ambiguity, other than when valves are placed in very abnormal configurations. We created the qualitative simulation to model the physical laboratory model, resulting in the aircraft engines being represented as tanks for example. The system involves a number of tanks, valves and pumps that allow fuel to be stored and transferred around the aircraft both to supply engines and to maintain aircraft trim during flight.

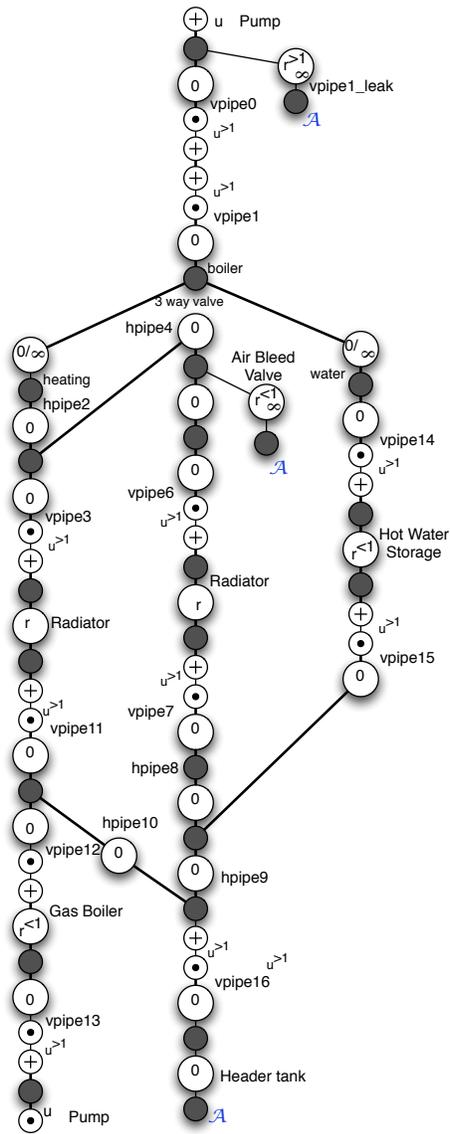

Figure 14: Flow model derived from Figure 12 (joints omitted)

The system is comprised of left and right fuel tanks situated in the aircraft wings (e.g. OC_WT_LH) and left and right auxiliary tanks (e.g. TK_AT_LH). The engines were represented in the physical test rig that was used for convenience, as tanks EH_LH and EH_RH. The wing tanks are connected to the engines via pumps (e.g. CP_FL_LH) and pressure sensors (e.g.

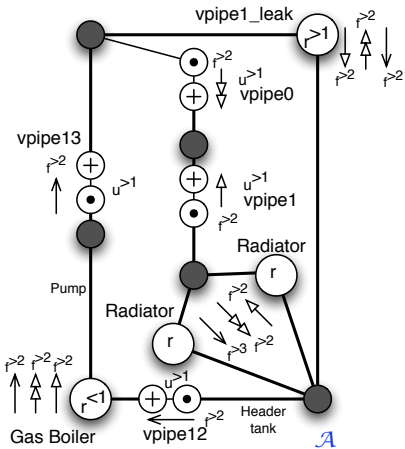

Figure 15: Simplified heating system with leak

PT_FL_LH) and flow metres (e.g. FT_FL_LH), which are also modelled as a tank to mimic the
hardware test rig that was used, with excess fuel being returned to the tank from which it
was drawn. Control of the fuel distribution is provided by four three-position valves (e.g.
TVL_FL...), which are slaved in pairs for the left and right subsystems. The basic operation
of the system is to supply fuel from the wing tanks to the corresponding engine by setting
all the valves in the normal position. These valves also allow fuel to be supplied from
the left tank to the right engine allowing both engines to be fed from one tank (crossover
operation). It is possible to feed both engines from opposite tanks if desired, although this
is not a normal operating mode. In addition fuel can be transferred between the wing tanks
and from the auxiliary tanks to the wing tanks, although it is not possible to return fuel to
the auxiliary tanks. Failure modes were provided for most component categories including
pipe and tank leaks, pump failures and stuck or leaking valve failures for every component
instance.

A portion of the resulting FMEA output is shown in Figure 17. The textual descrip-
tions are derived from the functional model (Bell et al., 2007; Price, 1998) and provide an
easily understood explanation of the fault effects and risk priority. Functions also inter-
pret exaggerated behaviours into human-friendly phrases, for example if the return line to
a tank has OM lower flow than nominal and the outflow is nominal, the virtual 'relative
level sensor' has a value lower than expected. Of course other faults may provide an absolute
qualitative value to the tank level sensor if it for example becomes empty (0) when part
filled ($l^{\triangleright 0}$) was expected. The consistency of the fully automated FMEA analysis allows
other automated tasks to be performed such as a diagnosability analysis (Snooke, 2009;
Snooke & Price, 2012). The qualitative analysis allows the entire FMEA to be regenerated
following system modification in a matter of seconds and only the differences to the system
effects are presented to the engineer as an incremental FMEA. This allows any unforeseen
implications of design changes to be easily detected.

The first row of Figure 17 describes the effects of a blocked fuel return pipe near the RH
engine in different operating modes of the system called 'steps' in the output. The main functional effect is a RH engine supply malfunction (too much fuel), and the effect that excess fuel is not returned to the tank. There is also an indication that the RH wing tank level might be lower than expected; of course this is a theoretical qualitative worst case.

| Boiler Pump | Radiator1 | Pipe0 | Pipe1 | Pipe3 | Pipe11 | Pipe12 | Pipe13 | Pipe16 | vpipe1_leak |
|---|---|---|---|---|---|---|---|---|---|
| on $F=f$ $\oplus=u$ | full | full $F=f$ $\oplus=u^{\triangleright 1}$ | full $F=-f$ $\oplus=u^{\triangleright 1}$ | full $F=-f$ $\oplus=u^{\triangleright 1}$ | full $F=f$ $\oplus=u^{\triangleright 1}$ | full $F=f$ $\oplus=u^{\triangleright 1}$ | full $F=f$ $\oplus=u^{\triangleright 1}$ | $F=f^{\triangleleft 2}$ $\oplus=u^{\triangleright 1}$ | $F=f^{\triangleleft 2}$ $\oplus=u^{\triangleright 1}$ |
| Close supply to header tank, and switch off boiler and pump. Flow non-determinism: $F=f$ because $E(\text{Pipe3})\leftrightarrow E(\text{Pipe11})$; $F=f$ because $E(\text{Pipe6})\leftrightarrow E(\text{Pipe7})$; Flow non-determinism: $F=f$ in $E(\text{Pipe1}) \leftrightarrow E(\text{Pipe0})$, $E(\text{Pipe12})$, $E(\text{Pipe13})$; **Resolve> Pipe3=Pipe11; Pipe6=Pipe7; Pipe1=Pipe0+Pipe13+Pipe12** | | | | | | | | | |
| off | empty | $F=-f^{\triangleleft 3}$ | $F=f^{\triangleleft 3}$ | $F=-f^{\triangleleft 3}$ | $F=f^{\triangleleft 3}$ | $F=f^{\triangleleft 2}$ | $F=f^{\triangleleft 2}$ | $\rightarrow$empty $\oplus=0$ | |
| | | | | | | $\rightarrow$empty $\oplus=0$ | | | |
| | | | | | | | $\rightarrow$empty $\oplus=0$ | | |
| | | | | $\rightarrow$empty $\oplus=0$ | $F=0$ | $F=0$ | | | $F=f^{\triangleleft 3}$ |
| Flow non-determinism: $F=-f^{\triangleleft 3}$ because $E(\text{Pipe3})\leftrightarrow E(\text{Pipe0})$; **Resolve> Pipe3>Pipe0;** | | | | | | | | | |
| | | $F=f^{\triangleleft 3}$ | $F=-f^{\triangleleft 3}$ | $F=f^{\triangleleft 3}$ | $F=-f^{\triangleleft 3}$ | | | | $F=-f^{\triangleleft 3}$ |
| Event non-determinism: Pipe0$\rightarrow$empty $\leftrightarrow$ Pipe11$\rightarrow$full **Resolve> Pipe0;** | | | | | | | | | |
| | | $\rightarrow$empty $\oplus=0$ | | | | | | | |
| Event non-determinism: Pipe1$\rightarrow$empty $\leftrightarrow$ Pipe11$\rightarrow$full **Resolve> Pipe11;** | | | | | | | | | |
| | | | | | $F=f^{\triangleleft 3}$ $\rightarrow$full $\oplus=u^{\triangleright 1}$ | | | | |
| | full | empty | full | full | full | empty | empty | empty | |

Table 5: Extract of simulation for leak fault in heating system 15

The second row deals with a fracture in the pipe just above the RH pump. The function effect is again that the engine supply failes in normal operating mode, but additionally we see that there is no flow at the flow transducer, and the RH wing tank level is higher than expected, although this is probably not the primary consideration, but could be used to indicate that fuel could be diverted to the remaining engine. This fault has a different effect in cross feed mode (fuel taken from the opposite wing tank), as in this case it is the LH tank level that is higher than expected due to the potential lack of returned fuel. Part of a RH valve fault is shown and we see that fuel is returned to the wrong tank when the LH engine is run.

The qualitative simulation has also been used to generate sets of symptoms that relate qualitative measurements, symptoms and failures (Snooke, 2009) in order to allow a diagnosability analysis to be performed with the aim of assisting sensor selection. The structure of the system has the greatest influence in these tasks and is exploited in a purely structural approach in related work (Rosich, 2012; Krysander & Frisk, 2008). For complex models using high order differential equations, dealing with sensing of individual components such as the valve example (Krysander & Frisk, 2008) the structure approach provides tractability. The system or product wide sensor placement analysis in early design investigation benefits in addition from the (abnormal) behaviour response between multiple interacting components in the presence of a fault, and is especially pertinent when only indirect sensing is possible. The less detailed qualitative approach allows both aspects to be used, while maintaining tractability. The comprehensive coverage of the system behaviour when linked with the functional states of the system allows automated fault isolation activities and this is the subject of another paper by one of the authors (Snooke & Price, 2012). The qualitative analysis is important for such tasks because it captures broad regions of similar system behaviour at a meaningful level of abstraction.

Multiple faults can be used in the simulation noting that some combinations of fault may produce qualitative ambiguity, representing critical tipping points of the system. For example in the fuel system each pump normally operates segments of the system partitioned by the valves. Multiple valve faults can result in the pumps working in opposition through a complex pipe topology and it is almost certain that numerical information is required to determine the actual flows because of the non-linearity abstracted into the qualitative component states. For an FMEA the resulting answer predicting several possible behaviours is reasonable, since if the highest risk effects are significant enough, they will be highlighted to the engineer for detailed analysis.

Single fault FMEA is the norm because of the effort involved in multiple fault effect determination. For the automated FMEA combinations of faults are feasible, although some selection of fault combinations is still usually necessary to alleviate the computational complexity $O(N^m)$ associated with exploring $m$ concurrent faults as has been previously discussed (Price & Taylor, 1997).

## 8. Conclusions

Qualitative simulation is a powerful modelling concept that can support a wide range of reasoning tasks. A range of electrical circuit design analysis tools have been based on this approach and the authors' electrical qualitative simulator known as MCIRQ is now in regular industrial use.

The structural and behavioural models are compositional and do not encode system functional information or make assumptions about their use. It is of course necessary to decide on the range and phenomena to be included in the modelling, and so a library of models is typically created for a specific application area (e.g. automotive electrical, aircraft fuel system, general plumbing etc), and this will include the set of qualitative variables of interest, their (suitably labelled) magnitudes and relevant component failure modes. The models are then reusable components available for other systems within the application area. The qualitative nature of the components makes them far less complex than numerical equivalents and this also makes them reusable within an application area – and possibly also to other application areas. The models should provide the major behaviours relevant to the objective of high-level reasoning about potential effects, and not detailed analysis of system performance, as has been demonstrated in the examples.

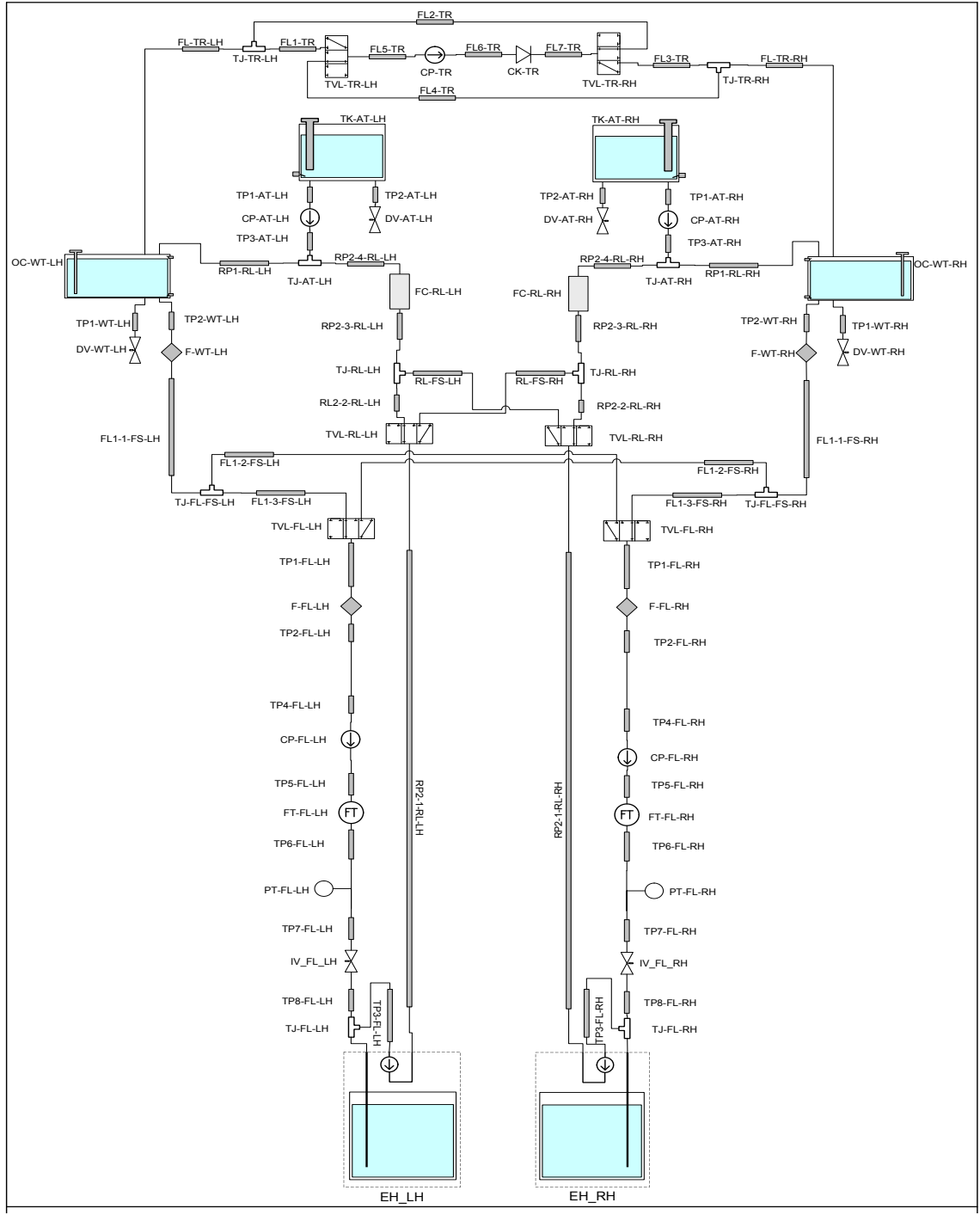

Figure 16: Example fuel system

The following table lists the results from the automatically generated FMEA

| Item | Behaviour | | | | Failure Causes | Sev | Det | Occ | RPN |
|---|---|---|---|---|---|---|---|---|---|
| **F1** | **At RH normal on (Step_6)** | | | | TP3_FL_RH - blocked. | 6 | 8 | | 48 |
| | Observable | Value | Function | Effects | S,D | | | | |
| | 'OC_WT_RH.tank_level' | lower than expected (normal level decrease expected) | 'engine_supply_right' failed | excess fuel not returned to tank | 6, 8 | | | | |
| | **At RH crossfeed on (Step_16)** | | | | | | | | |
| | Observable | Value | Function | Effects | S,D | | | | |
| | 'OC_WT_LH.tank_level' | lower than expected (normal level decrease expected) | 'engine_crossfeed_to_right' failed | excess fuel not returned to tank | 6, 8 | | | | |
| **F2** | **At RH normal on (Step_6)** | | | | TP2_FL_RH - fracture. | 8 | 8 | | 64 |
| | Observable | Value | Function | Effects | S,D | | | | |
| | 'FT_FL_RH.flow' | none (normal expected) | 'engine_supply_right' failed | possible engine malfunction | 8, 8 | | | | |
| | 'OC_WT_RH.tank_level' | higher than expected (normal level decrease expected) | | | | | | | |
| | **At RH crossfeed on (Step_16)** | | | | | | | | |
| | Observable | Value | Function | Effects | S,D | | | | |
| | 'FT_FL_RH.flow' | none (normal expected) | 'engine_crossfeed_to_right' failed | possible engine malfunction | 8, 8 | | | | |
| | 'OC_WT_LH.tank_level' | higher than expected (normal level decrease expected) | | | | | | | |
| **F3** | **At Start - PRIOR TO FIRST EXTERNAL CHANGE (Step_1)** | | | | TVL_RL_RH - stuck_crossover. | 6 | 8 | | 48 |
| | Observable | Value | | | | | | | |
| | 'TVL_RL_RH.position_tellback' | crossover (isolation expected) | | | | | | | |
| | **At LH normal set (Step_2)** | | | | | | | | |
| | Observable | Value | | | | | | | |
| | 'ALL' | as previous step | | | | | | | |
| | **At LH normal on (Step_3)** | | | | | | | | |
| | Observable | Value | Function | Effects | S,D | | | | |
| | 'ALL' | as previous step | 'engine_crossfeed_to_right' unexpected | fuel returned to left tank when not expected | 1, 8 | | | | |
| | **At LH normal off (Step_4)** | | | | | | | | |

Figure 17: Portion of FMEA output for fuel system

This paper makes two notable contributions. Firstly, it provides an improved circuit reasoning algorithm that gives a complete solution for all possible circuit topologies. This is achieved by solving the problem of non series/parallel reducible circuits. Also the restriction on single sources is removed and the resultant simulator is called (M²CIRQ), for Multiple source MCIRQ.

Secondly, the qualitative network modelling method is placed in the context of a modelling ontology based on separating global and local behaviour based on power flow. The component models for fluid systems involve more aspects than electrical circuits and the paper introduces several additional fundamental concepts necessary at the global level for fluid flow modelling, including a distinguished zero node and propagation of substances through a network. These techniques are illustrated by modelling a range of common fluid flow components for simulation.

The ability of the QR to make predictions across multiple system states and operating modes is complementary to other techniques. For example fault tree analysis for diagnosis of a very similar fuel system to the example presented in Section 7 has been performed (Hurdle, Bartletta, & Andrews, 2009) and uses pattern recognition to deal with multiple states. The production of the fault trees and scenario (state) identification is a labour intensive manually performed process and is described in exactly the same qualitative terms as produced by QR behaviour prediction such as 'high flow'. It is therefore likely that accuracy and coverage of the FTA might be improved with a reduction in effort by using the QR behaviour predictions instead of manual effort.

Many software tools have been developed to perform a variety of design analysis for electrical systems, as mentioned in the introduction to this paper. By substituting the enhanced simulator into these tools the same analyses can now be performed on other types of complex topology systems and multiple domain systems where qualitative behaviours can be used to answer failure mode questions.

## 8.1 Acknowledgments

This work was supported by Aberystwyth University, the Welsh Assembly Government, BAE Systems and the DTI ASTRAEA Programme. We also thank the anonymous reviewers for their helpful comments.